\documentclass[runningheads]{llncs}
\usepackage[T1]{fontenc}
\usepackage{graphicx}

\usepackage{hyperref}
\usepackage{url}
\usepackage{subcaption}
\usepackage{algorithm}
\usepackage{algorithmic}
\usepackage{multirow}
\usepackage{placeins}
\usepackage{booktabs}
\usepackage{amssymb}
\usepackage[inline]{enumitem}
\begin{document}
\pagestyle{plain}
\title{Beyond Explanation: Debugging Medical Imaging Models via Concept Intervention}
\titlerunning{Debugging Medical Models via Concept Intervention}

\author{Samrajya Thapa\inst{1}\thanks{Corresponding author.} \and
Daniel J. Quest\inst{2} \and 
Timothy L. Kline\inst{2} \and 
Carrie L. Langstraat\inst{2} \and 
Emanuel C. Trabuco\inst{2} \and
Wei Le\inst{1}}


\authorrunning{S. Thapa et al.}

\institute{Iowa State University, Ames IA, USA \and
Mayo Clinic, Rochester MN, USA\\
\email{svthapa@iastate.edu} \and
\email{weile@iastate.edu}}


\maketitle              

\begin{abstract}

Medical imaging models often operate as black boxes, limiting interpretability and systematic debugging. We introduce an easy-to-use, plug-and-play framework for concept-based interpretation and model refinement. By aligning a single-modality encoder \cite{moayeri2023text} to BioMedCLIP \cite{zhang2023biomedclip}, we construct a Concept Bottleneck Model (CBM) that enables concept-level interventions. These interventions allow us to isolate causal versus spuriously correlated concepts, validate insights with domain experts, and generate counterfactual samples for targeted fine-tuning. We evaluate our framework on a Mayo Clinic ultrasound dataset and the CheXpert 5x200 chest X-ray dataset \cite{cohen2022torchxrayvision}. Results demonstrate that concept intervention enables reliable model diagnosis while maintaining, and occasionally improving predictive performance via guided fine-tuning. Our findings highlight the practical value of this framework for controlled, interpretable refinement of clinical deep learning models.

\end{abstract}
\section{Introduction}

Deep learning models achieve strong performance across diverse medical imaging tasks, including disease classification, segmentation, and prognosis. Model Interpretability is key in establishing trust between the practicing clinitian and AI models. We need models that can describe what they segment or classify in understandable text. Most modern image encoders operate as black boxes, producing predictions without explicit links to clinically meaningful abstractions, which complicates trust, validation, and systematic error analysis by domain experts in clinical environments. 

Traditional post-hoc interpretability approaches \cite{patricio2023explainable,bhati2024survey} predominantly rely on visualization techniques such as saliency maps or gradient-based attribution. While these methods highlight regions of interest, they lack semantic grounding, exhibit high sensitivity to perturbations, and fail to align with clinical reasoning \cite{adebayo2018sanity}. Clinicians typically interpret medical images through structured, human-understandable concepts, such as radiographic findings, rather than pixel-level activations. Bridging this gap remains an open challenge.
Concept Bottleneck Models (CBMs) \cite{koh2020concept} offer a promising alternative by explicitly reasoning over predefined concepts. However, most CBM approaches require dense concept-level supervision and end-to-end training \cite{patricio2023coherent}, limiting their applicability to pre-trained, frozen, or deployed models. Concurrently, vision–language (VL) foundation models like CLIP \cite{radford2021learning} demonstrate strong semantic alignment between images and text, yet their clinical utility has largely remained confined to passive zero-shot classification or retrieval, rather than enabling fine-grained debugging of task-specific models. 

In this work, we propose a plug-and-play, concept-based fine-tuning framework that leverages large vision–language models as a shared semantic representation space to interpret and debug single-modality medical imaging models. Following recent insights that a linear mapping is sufficient to align disparate representation spaces \cite{moayeri2023text,oikarinen2022clip}, we project a task-specific image encoder into the BioMedCLIP \cite{zhang2023biomedclip} embedding space. This projection enables semantic grounding without requiring intrinsic textual supervision or end-to-end retraining. On top of this aligned representation, we construct a CBM using concepts derived from radiology reports or structured annotations. Crucially, we leverage gradient-guided concept-level interventions to turn interpretability into an active mechanism for model optimization, allowing us to systematically isolate missing, weak, or spurious concepts. 

Our contributions are summarized as follows: (i)~\textbf{Modular Architecture:} We introduce a plug-in concept bottleneck model (CBM) enabling concept-level interpretability for off-the-shelf single-encoder medical models via linear alignment to BioMedCLIP; (ii)~\textbf{Expert Validation:} We validate the clinical utility of our concept-based explanations with domain experts, demonstrating tight alignment with clinical reasoning; (iii)~\textbf{Causal Discovery:} We leverage intervenable CBMs for counterfactual concept interventions to systematically isolate causal concepts from spurious features; and (iv)~\textbf{Intervention-Guided Refinement:} We utilize our intervention strategy to apply counterfactual concept perturbations to held-out data to fine-tune the classification head and boost downstream predictive performance. 

We evaluate this framework on the Mayo Clinic ultrasound and CheXpert 5$\times$200 chest X-ray datasets. With this modular plug-in tool, we are able to identify spurious and causal correlations in a highly data-efficient manner, using clinical concepts to train models to dial into the correct diagnostic features within the image and more closely align human medical learning with traditional machine learning. While human clinicians excel at learning a great deal from a small amount of data, we argue that with a Concept Bottleneck Model (CBM), machines can do so as well. Crucially, our framework achieves this paradigm shift while maintaining baseline performance and systematically boosting downstream capabilities (2--3\%) through our targeted intervention strategy.

\section{Related Work}

\paragraph{\textbf{Feature Attribution vs. Concept Interpretability}}

While widely adopted, literature demonstrates that these pixel-level heatmaps inherently suffer from notable vulnerabilities, frequently offering purely qualitative visualizations that lack structured semantic grounding while exhibiting high sensitivity to network architectures and minor input perturbations \cite{adebayo2018sanity,smilkov2017smoothgrad}. Concept-based paradigms fundamentally deviate from this approach; instead of attempting to interpret ungrounded pixel-level features post-hoc, they bypass these established visualization constraints by routing downstream predictions directly through an intermediate layer of explicit, human-understandable semantic representations \cite{koh2020concept}.

\paragraph{\textbf{Vision-Language Alignment and Post-Hoc CBMs}} 
Standard CBMs typically require dense, manual concept annotations and end-to-end retraining, limiting their utility for pre-trained clinical models \cite{patricio2023coherent}. To bypass this bottleneck, recent post-hoc or label-free CBMs leverage the joint semantic space of vision–language models like CLIP \cite{radford2021learning} and BioMedCLIP \cite{zhang2023biomedclip}. These approaches project single-modality image representations onto textual concept spaces without complete model retraining \cite{moayeri2023text}. We build on this paradigm, demonstrating that a lightweight affine mapping is sufficient to anchor complex, single-modality medical encoders within a clinically grounded concept space.

\paragraph{\textbf{Concept Intervention and Model Refinement}} 
Test-time concept interventions allow researchers to simulate counterfactual states, probe model robustness, and analyze causal failure modes \cite{yeh2020completeness}, \cite{shin2023closer}, \cite{koh2020concept}. However, existing frameworks typically assume access to full concept labels during training or couple concept prediction tightly with primary task learning. Utilizing expert-guided concept interventions specifically to debug and fine-tune frozen, single-modality medical models remains largely underexplored. Our work bridges this gap by transforming test-time interventions into an active mechanism for targeted latent data augmentation and classification head refinement.
\section{Methodology}

\begin{figure}[t]
    \centering
    \includegraphics[width=\linewidth]{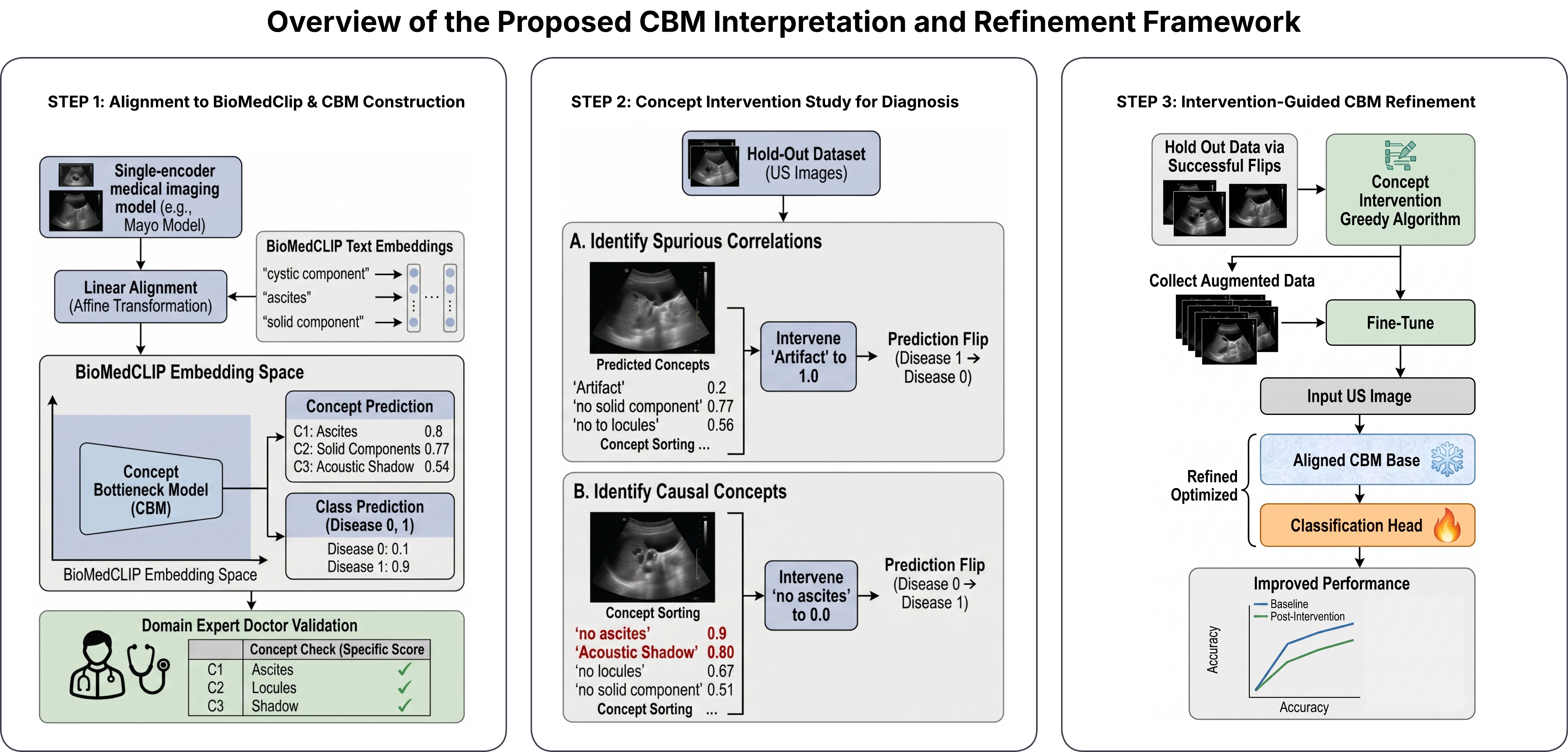}
    \caption{Overview of the proposed framework. A single-modality medical imaging model is linearly aligned to the BioMedCLIP embedding space to enable concept-based interpretability via a concept bottleneck model (CBM). Concept interventions on held-out data identify causal versus spurious concepts through prediction flips. These interventions are then used to augment data and fine-tune the CBM classification head, improving predictive performance.}
    \label{fig:overview}
\end{figure}

Our methodology enables concept-based interpretability and controlled refinement of single-modality medical imaging models by aligning their latent representations to a shared vision--language semantic space. The framework consists of three core phases: (i)~\textbf{Post-Hoc Mapping} extracts features from a task-specific clinical image encoder and linearly aligns its representation space to the BioMedCLIP embedding space; (ii)~\textbf{Concept Bottleneck Construction} generates semantically grounded concept predictions using textual abstractions derived from clinical knowledge; and (iii)~\textbf{Counterfactual Intervention} leverages expert-guided concept perturbations to isolate causal factors, flag spurious concepts, and drive targeted counterfactual data augmentation for downstream classification head fine-tuning. Overview of the pipeline is illustrated in Figure~\ref{fig:overview}
\subsection{Post-Hoc Alignment and CBM Construction}

We begin with a pre-trained, off-the-shelf single-modality medical imaging model that operates purely on images and lacks intrinsic textual supervision. To obtain a stable and semantically rich representation, we extract image features from an intermediate layer prior to the final classification layer. Specifically, features are isolated after global average pooling and before the final nonlinearity, as this representation preserves linear structure while retaining highly discriminative information. These extracted features serve as fixed image embeddings.

To enable semantic grounding, we align the feature space of this single-modality model to the embedding space of BioMedCLIP \cite{zhang2023biomedclip}, a large-scale vision--language model pre-trained on approximately 15 million medical image--report pairs. We explicitly note that foundation models like BioMedCLIP are fundamentally pre-trained on public scientific literature rather than native, institutional Electronic Health Record (EHR) systems. Furthermore, clinical radiology notes typically describe global, multi-view findings across an entire study or scan volume rather than localizing concepts to specific 2D images or individual slices. While a production-level deployment would ideally be trained natively on paired local EHR data, this work serves as an intentional proof-of-concept to establish the baseline feasibility of post-hoc cross-model alignment under these broader domain constraints. Following prior work on cross-model alignment \cite{moayeri2023text}, we learn an affine transformation that maps the single-modality image features directly into this shared BioMedCLIP embedding space. 

\subsection{Counterfactual Concept Intervention and Diagnosis}
We formulate model validation as a problem of counterfactual concept-level intervention to cleanly isolate spurious correlations from informative clinical features. Formally, an intervention operator $\mathcal{I}(\hat{\mathbf{c}}, \mathcal{M})$ overwrites a targeted subset of predicted concept activations $\hat{\mathbf{c}}$ with alternative binary states specified by an intervention map $\mathcal{M}$, generating a counterfactual concept vector $\tilde{\mathbf{c}}$. 

To efficiently identify the minimal concept set necessary to alter the model's decision, we implement the gradient-guided greedy strategy detailed in Algorithm~\ref{alg:greedy_intervention} (inspired by \cite{shin2023closer}). This protocol prioritizes candidate interventions by ranking concepts according to their absolute gradient magnitude with respect to the target class logit, $\nabla_{\hat{\mathbf{c}}} g_{y^*}(\hat{\mathbf{c}})$.

\begin{algorithm}[h]
\caption{Greedy Concept-Level Intervention}
\label{alg:greedy_intervention}
\footnotesize
\begin{algorithmic}[1]
\REQUIRE Concept activations $\hat{\mathbf{c}} \in \mathbb{R}^m$, classification head $g(\cdot)$, target label $y^*$, intervention operator $\mathcal{I}(\cdot)$, maximum concept search depth $K$
\ENSURE Minimal set of concept interventions $\mathcal{M}$ that flips the prediction to $y^*$

\STATE Compute original baseline prediction $\hat{y} = \arg\max g(\hat{\mathbf{c}})$
\IF{$\hat{y} = y^*$}
    \RETURN $\emptyset$ \COMMENT{Target prediction already satisfied}
\ENDIF

\STATE Compute gradient vector $\mathbf{G} = \nabla_{\hat{\mathbf{c}}} g_{y^*}(\hat{\mathbf{c}})$ of the target class logit with respect to concept activations
\STATE Rank concepts $\{c_1, \dots, c_m\}$ by descending absolute gradient magnitude of $\mathbf{G}$

\FOR{$k = 1$ to $K$}
    \STATE Isolate top-$k$ ranked concepts $\mathcal{C}_k$
    \FOR{each binary counterfactual assignment $\mathbf{v} \in \{0,1\}^k$}
        \STATE Construct intervention map $\mathcal{M} = \{(c_i \leftarrow v_i) \mid c_i \in \mathcal{C}_k\}$
        \STATE Apply intervention to synthesize counterfactual: $\tilde{\mathbf{c}} = \mathcal{I}(\hat{\mathbf{c}}, \mathcal{M})$
        \STATE Compute counterfactual prediction $\tilde{y} = \arg\max g(\tilde{\mathbf{c}})$
        \IF{$\tilde{y} = y^*$}
            \RETURN $\mathcal{M}$ \COMMENT{Minimal flipping set discovered}
        \ENDIF
    \ENDFOR
\ENDFOR

\STATE \RETURN $\emptyset$ \COMMENT{No successful intervention set found within depth $K$}
\end{algorithmic}
\end{algorithm}

Interventions are applied sequentially to the top-$k$ concepts by pushing their activations toward binary extremes. The smallest intervention set that successfully induces a prediction flip provides a concise counterfactual explanation of the model’s decision boundary. 

\subsection{Intervention-Guided Refinement}
We leverage successful prediction flips from the diagnostic phase to drive targeted model refinement. When Algorithm~\ref{alg:greedy_intervention} identifies an intervention map $\mathcal{M}$ that corrects an erroneous baseline prediction, the resulting counterfactual concept vector $\tilde{\mathbf{c}}$ is aggregated into a concept-guided counterfactual augmentation dataset $\mathcal{D}_{aug} = \{(\tilde{\mathbf{c}}_n, y^*_n)\}$. Operating entirely within the latent bottleneck layer avoids computationally expensive or ungrounded pixel-level image synthesis while generating targeted training signals that expose decision-boundary vulnerabilities.To prevent representation collapse and preserve feature robustness, the image encoder $f_s(\cdot)$ and linear alignment matrix $W$ remain frozen. Fine-tuning is strictly restricted to the linear task classification head $g(\cdot)$. Optimizing $g(\cdot)$ over $\mathcal{D}_{aug}$ explicitly penalizes reliance on spurious artifacts and reinforces sensitivity to causal clinical markers, improving diagnostic accuracy without sacrificing baseline generalization.

\subsection{Concept Selection and Expert Validation}
For the Mayo dataset, we derived 34 concepts from clinical metadata, encompassing both categorical and numerical structural features (e.g., ascites, locules, solid/cystic components, papillary projections). Numerical features were statistically binned through training data analysis, and clinical domain experts were consulted for final concept curation. For CheXpert, we extracted 10 discriminative concepts from paired radiology reports, establishing ground truths via regex matching with strict negation detection to ensure genuine concept presence. 

To validate this concept layer (Figure \ref{fig:multimodal_intervention}), Mayo predictions and counterfactual interventions were evaluated by domain experts on a held-out, out-of-distribution clinical cohort, while CheXpert predictions were cross-checked directly against findings in the original text reports.
\begin{table}[t]
\centering
\caption{Performance evaluation on held-out Mayo Clinic test splits comparing the baseline ensemble, pre-intervention, and post-intervention fine-tuned CBM.}
\label{tab:mayo_cbm_combined}
\begin{tabular}{lcccc}
\hline
\textbf{Model Configuration} & \textbf{Test Set} & \textbf{Accuracy} & \textbf{F1} & \textbf{AUROC} \\
\hline
\multirow{2}{*}{Mayo Ensemble (Baseline)}   & Test Set 1 & 0.890 & 0.880 & 0.920 \\
                                            & Test Set 2 & 0.730 & 0.740 & 0.840 \\
\hline
\multirow{2}{*}{CBM (Pre-Intervention)}     & Test Set 1 & 0.850 & 0.850 & 0.916 \\
                                            & Test Set 2 & 0.740 & 0.740 & 0.839 \\
\hline
\multirow{2}{*}{CBM (Post-Intervention)}    & Test Set 1 & 0.850 & 0.850 & \textbf{0.917} \\
                                            & Test Set 2 & \textbf{0.753} & \textbf{0.749} & 0.839 \\
\hline
\end{tabular}
\end{table}

\begin{table}[t]
\centering
\caption{CheXpert per-label performance comparison between TorchXrayVision baseline and our CBM framework before and after intervention-based fine-tuning.}
\label{tab:chexpert}
\resizebox{\linewidth}{!}{%
\setlength{\tabcolsep}{2pt} 
\begin{tabular}{lcccccccccccc}
\toprule
\multirow{2}{*}{\textbf{Pathology Class}} 
& \multicolumn{4}{c}{\textbf{TorchXrayVision Baseline}}
& \multicolumn{4}{c}{\textbf{CBM (Pre-Intervention)}} 
& \multicolumn{4}{c}{\textbf{CBM (Post-Intervention)}} \\
\cmidrule(r){2-5} \cmidrule(lr){6-9} \cmidrule(l){10-13}
& Prec & Rec & F1 & AUC & Prec & Rec & F1 & AUC & Prec & Rec & F1 & AUC \\
\midrule
Cardiomegaly   & 0.63 & \textbf{0.68} & 0.65 & \textbf{0.86} & \textbf{0.73} & 0.58 & 0.65 & 0.85 & \textbf{0.73} & 0.58 & 0.65 & 0.85 \\
Consolidation  & \textbf{0.50} & 0.05 & 0.09 & 0.71 & 0.39 & \textbf{0.71} & 0.51 & 0.69 & 0.43 & \textbf{0.71} & \textbf{0.54} & \textbf{0.75} \\
Edema          & 0.43 & 0.21 & 0.28 & 0.70 & \textbf{0.48} & 0.45 & 0.47 & \textbf{0.83} & \textbf{0.48} & \textbf{0.50} & \textbf{0.49} & 0.80 \\
Effusion       & 0.35 & \textbf{0.66} & 0.46 & 0.77 & \textbf{0.73} & 0.61 & \textbf{0.67} & 0.89 & \textbf{0.73} & 0.61 & \textbf{0.67} & \textbf{0.90} \\
Atelectasis    & 0.43 & \textbf{0.63} & 0.51 & 0.76 & 0.91 & 0.50 & 0.65 & 0.84 & \textbf{0.92} & 0.55 & \textbf{0.69} & \textbf{0.90} \\
\midrule
\textbf{Macro Avg}    
& 0.47 & 0.45 & 0.40 & 0.76 
& 0.65 & 0.57 & 0.59 & 0.82 
& \textbf{0.66} & \textbf{0.59} & \textbf{0.61} & \textbf{0.84} \\
\textbf{Weighted Avg} 
& 0.47 & 0.45 & 0.40 & —    
& 0.64 & 0.57 & 0.58 & —    
& \textbf{0.65} & \textbf{0.59} & \textbf{0.60} & — \\
\bottomrule
\end{tabular}%
}
\end{table}

\begin{table}[t]
\centering
\caption{Overall performance comparison across models on CheXpert 5×200.}
\label{tab:overall_perf}
\begin{tabular}{lcccc}
\hline
\textbf{Model} & \textbf{Accuracy} & \textbf{Macro F1} & \textbf{AUROC}  & \textbf{Interpretability} \\
\hline
CLIP Zero-shot & 33.9 & 27.4 & 62.3 & No \\
Aligned Model (Zero-shot) & 33.3 & 24.5 & 64.1 & Limited \\
X-ray Vision (Linear Probe) & 53.7 & 52.0 & 82.1 & No \\
CBM (Pre-intervention) & 57.0 & 59.0 & 82.0 & Yes \\
CBM (Post-intervention) & \textbf{59.0} & \textbf{61.0} & \textbf{84.0} & Yes \\
\hline
\end{tabular}
\end{table}

\section{Results}
We evaluate our framework on a clinical ovarian ultrasound dataset from Mayo Clinic (525 samples; 361 train, 164 test) alongside an additional out-of-distribution validation set of 644 samples. To demonstrate domain generalization, we further evaluate on CheXpert 5$\times$200, a balanced multi-label benchmark consisting of 1,000 chest X-rays across five clinical findings, utilizing a pre-trained DenseNet backbone from TorchXrayVision \cite{cohen2022torchxrayvision} as our base model. 

\paragraph{\textbf{CLIP Alignment Sanity Check}}
To address our first claim, we verify that mapping features to the BioMedCLIP space preserves representation quality. On the CheXpert benchmark in Table \ref{tab:overall_perf}, the aligned zero-shot model achieves 33.3\% accuracy compared to the CLIP baseline of 33.9\%, while demonstrating a minor 1.8\% improvement in AUROC (64.1\% vs. 62.3\%). This demonstrates that our learned projection layer retains downstream representation quality within a 1\% absolute performance margin, serving as a vital sanity check that we preserve clinical feature discrimination without intensive end-to-end backbone retraining. 

\paragraph{\textbf{Concept Bottleneck and Intervention Performance}}
Table~\ref{tab:mayo_cbm_combined} tracks the progression of performance metrics across the baseline ensemble and our CBM variants on two held-out ultrasound test splits. On the in-distribution Test Set 1, routing predictions through an explicit concept layer preserves the high task accuracy of the uninterpretable baseline, while post-intervention fine-tuning yields a marginal edge in AUROC. On the more complex, out-of-distribution Test Set 2, the pre-intervention CBM performs consistently with baseline trends, whereas targeted counterfactual refinement delivers noticeable improvements in both accuracy and F1-score. These results verify that our framework optimizes downstream decision boundaries without risking performance degradation or representation collapse.

\paragraph{\textbf{Additional Dataset Study}}
Table~\ref{tab:chexpert} presents fine-grained per-label performance on the CheXpert 5x200, contrasting the baseline TorchXrayVision model against our framework. Compared to the uninterpretable baseline, the pre intervention CBM establishes a significantly more balanced trade-off between precision and recall, correcting severe baseline sensitivity flaws on pathologies such as Consolidation. Post-intervention counterfactual fine-tuning delivers further across-the-board gains, improving macro-averaged recall, F1-score, and AUROC.

Finally, Table~\ref{tab:overall_perf} compiles the global performance metrics across all evaluated frameworks on the CheXpert benchmark. While standard zero-shot and linear probing settings provide competitive baselines, our intervention-guided CBM achieves the highest overall accuracy and Macro F1 while uniquely providing intrinsic model interpretability.

\begin{figure*}[!t]
    \centering
    \includegraphics[width=\linewidth]{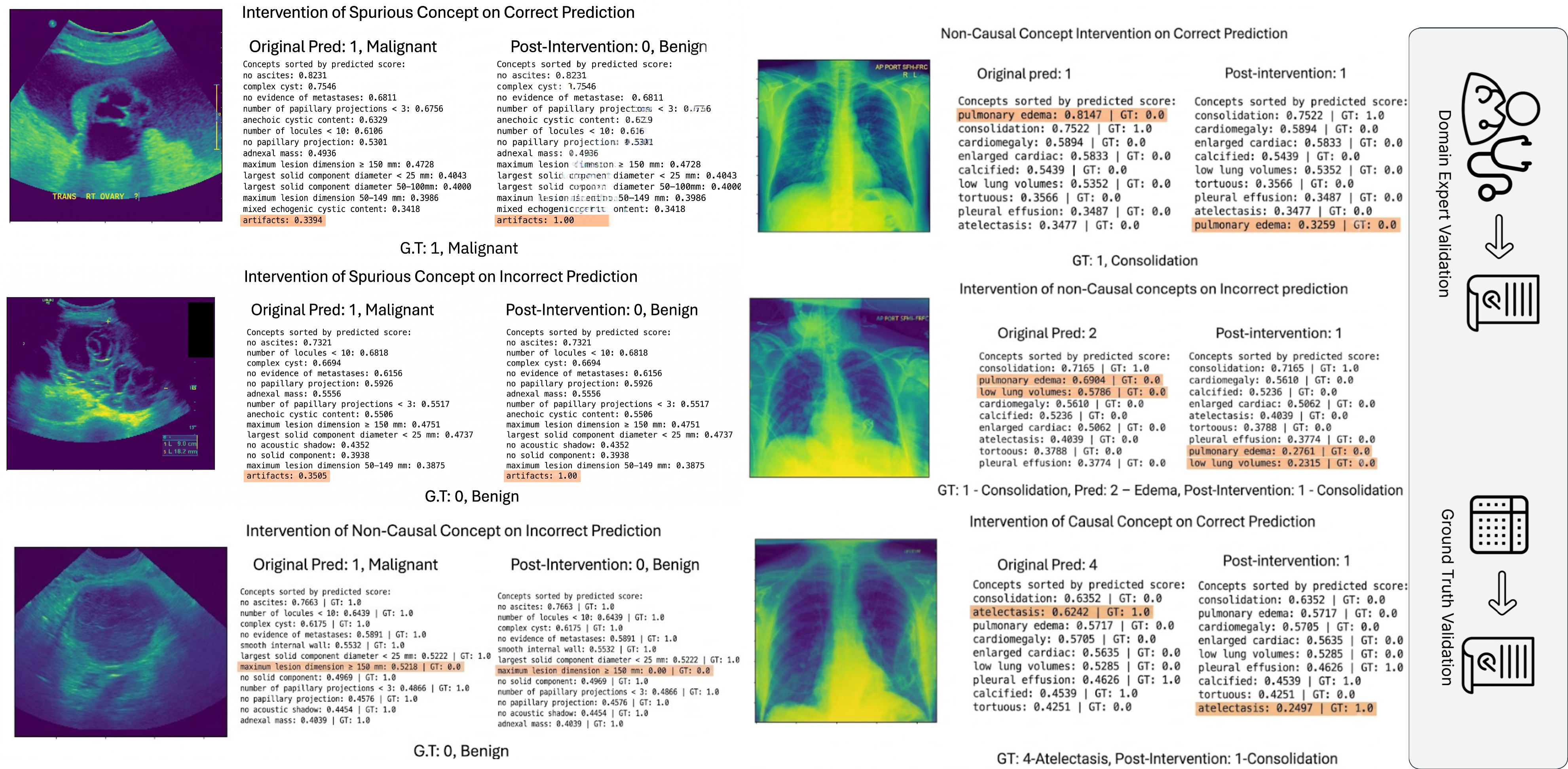}
    \vspace{-3mm}
    \caption{Interpretation and intervention case studies across ultrasound (Mayo Clinic) and chest X-ray (CheXpert) modalities}
    \label{fig:multimodal_intervention}
\end{figure*}

\paragraph{\textbf{Concept-Based Explanation and Intervention}}
We qualitatively evaluate our framework's debugging utility via case studies from Mayo Clinic ultrasound and CheXpert $5\times200$ datasets (Figure~\ref{fig:multimodal_intervention}). Unlike traditional image-only models that can overfit to non-clinical artifacts like scanner types or text annotations, our architecture isolates these variables to allow test-time counterfactual interventions. Specifically, adjusting spurious concept weights rectifies false predictions on ovarian masses (left column) and resolves multi-label ambiguities in chest X-rays (right column). This parameter-free auditing mechanism demonstrates how concept-level interventions compensate for imperfect datasets, enabling high-quality clinical modeling with superior sample efficiency and robust insulation from spurious features.

\section{Conclusion}
We present a plug-in framework for concept-based interpretability and refinement of off-the-shelf medical deep learning models \cite{koh2020concept}. By linearly mapping a pre-trained encoder to the BioMedCLIP embedding space, our approach constructs an intervenable Concept Bottleneck Model (CBM) without architectural modifications. Beyond passive explanation, gradient-guided concept interventions enable domain experts to perform counterfactual debugging and targeted classification head fine-tuning, directly correcting systematic failure modes to enhance trust in clinical deployments.


A key limitation is the spatial and structural disconnect between global clinical narratives and specific image frames. While rich semantic concepts are natively documented within radiology notes at the time of diagnosis, these clinical summaries describe entire multi-view studies globally and lack granular, instance-level association with individual images or slices. Consequently, rather than relying on uncalibrated, off-the-shelf foundation spaces, future research will focus on developing weakly-supervised alignment techniques to dynamically map text-mined concepts from native clinical reports directly to corresponding image volumes \cite{bhalla2024interpreting}, mitigating data alignment bottlenecks across diverse clinical environments.

\paragraph{\textbf{Prospects of Application:}} This framework can be deployed as an auditing plug-in for legacy clinical workflows. In interactive environments, it enables real-time, human-in-the-loop decision debugging, allowing clinicians to dynamically verify model reasoning, suppress misleading artifacts, and safely refine diagnostic heads at the point of care without end-to-end retraining.

\subsubsection{Acknowledgements}
The authors gratefully acknowledge Dr.~Luigi De Vitis for his clinical guidance in this project, as well as AI researchers Dr.~Adriana Gregory and Dr.~Mrinal Dhar for their technical insights. Finally, special thanks to Operations Manager Joseph Fulton for facilitating this collaboration and establishing the operational framework for this research.

\subsubsection{\discintname}
The authors have no competing interests to declare that are relevant to the content of this article.
%
%
%
\bibliographystyle{splncs04}
\bibliography{miccai_conference}

\end{document}